\documentclass[conference]{IEEEtran}
\IEEEoverridecommandlockouts
\usepackage{cite}
\usepackage{amsmath,amssymb,amsfonts}
\usepackage{algorithmic}
\usepackage{graphicx}
\usepackage{textcomp}
\usepackage{xcolor}
\usepackage{booktabs} % For formal tables
\usepackage{caption}
\usepackage{wrapfig}
\usepackage{subfig}
\usepackage{multirow}
\usepackage{url}
\usepackage[normalem]{ulem}
\def\BibTeX{{\rm B\kern-.05em{\sc i\kern-.025em b}\kern-.08em
    T\kern-.1667em\lower.7ex\hbox{E}\kern-.125emX}}
\begin{document}

\title{Automatic Compiler Based FPGA Accelerator \\ for CNN Training
  \vskip -0.15in
%{\footnotesize \textsuperscript{*}Note: Sub-titles are not captured in Xplore and
%should not be used}
\thanks{The authors would like to thank Intel Corporation for supporting and funding this research work.
This work was also partially supported by NSF grant 1652866 and C-BRIC, one of six centers in JUMP, a SRC program sponsored by DARPA.}
}

%added by shreyas
\graphicspath {}

\author{
    \IEEEauthorblockN{Shreyas Kolala Venkataramanaiah, Yufei Ma, Shihui Yin, Eriko Nurvithadhi\IEEEauthorrefmark{1}, Aravind Dasu\IEEEauthorrefmark{2}, Yu Cao, Jae-sun Seo}
    \IEEEauthorblockA{School of Electrical, Computer and Energy Engineering, Arizona State University, Tempe, AZ, USA}
    \IEEEauthorblockA{\IEEEauthorrefmark{1}Intel Labs, Intel Corporation, Hillsboro, OR, USA}
    \IEEEauthorblockA{\IEEEauthorrefmark{2}Programmable Solutions Group, Intel Corporation, San Jose, CA, USA}
    
    Email: skvenka5@asu.edu
}

\maketitle

\begin{abstract}
Training of convolutional neural networks (CNNs) on embedded platforms to support on-device learning is earning vital importance in recent days. 
%Many FPGA accelerators have been proposed for CNN inference, but limited work has been reported on implementing training algorithms. %Prior works on CNN training either use FPGAs as CNN inference accelerator and update weights on CPU %\cite{zhao2016f}  
%or focus on dedicated hardware for weight update. %\cite{choi2018trainware}. 
%However, these works do not provide a configurable FPGA training accelerator to implement Forward Pass (FP), Backward Pass (BP) and Weight Update (WU). 
%In particular, prior works do not present an end-to-end FPGA training accelerator to implement Forward Pass (FP), Backward Pass (BP) and Weight Update (WU). 
Designing flexible training hardware is much more challenging than inference hardware, due to design complexity and large computation/memory requirement. %, a high degree of design complexity and vast memory requirement. 
In this work, %extends the work CNN inference accelerator \cite{ma2017automatic,ma2016scalable,ma2017optimizing} and 
we present an 
automatic compiler based FPGA accelerator with 16-bit fixed-point precision for complete CNN training, including Forward Pass (FP), Backward Pass (BP) and Weight Update (WU). 
%using stochastic gradient descent with momentum based weight update using 16-bit fixed-point precision.  
We implemented an optimized RTL library to perform training-specific tasks, % including stochastic gradient descent and momentum based weight update,  
%weight gradient computation and BP convolutions using flipped kernels, 
and developed an RTL compiler to automatically generate FPGA-synthesizable RTL based on user-defined constraints.
%Operations in each iteration (FP, BP, and WU) are categorized into different layers and processed sequentially based on execution schedule. 
We present a new cyclic weight storage/access scheme for on-chip BRAM and off-chip DRAM to efficiently implement non-transpose and transpose operations during FP and BP phases, respectively.
Representative CNNs for CIFAR-10 dataset are implemented and trained on Intel Stratix 10 GX FPGA using proposed hardware architecture, demonstrating up to 479 GOPS performance. 

%\alpha\beta^n\times(Q_i - Q_a)

\end{abstract}

\begin{IEEEkeywords}
Convolution neural networks, neural network training, back-propagation, hardware accelerator, FPGA
\end{IEEEkeywords}

\section{Introduction}
%As one of the most successful deep learning algorithms, 
CNNs have shown tremendous performance in many practical tasks including computer vision~\cite{hu2018squeeze} and speech recognition~\cite{zhang2016interspeech}. 
Deep CNNs %with complex training algorithms are used to 
achieve high accuracy on large datasets, but an enormous amount of computation is required for training such networks. %computational complexity. 
To support the high computation requirement, training tasks have been typically performed on datacenters with high-end GPUs. Nowadays, training on resource-constrained platforms is becoming more crucial for training networks with each user's private data.
However, executing computation-/memory-intensive training tasks on hardware platforms with power and resource constraints become very challenging. This gives an opportunity to map these algorithms on FPGAs, which provide high configurability and power-efficiency compared to those of GPUs. They also provide a large volume of off-chip memory (DRAM) and shorter design time when compared to ASIC designs. 

For CNN inference tasks, a number of FPGA accelerators have been proposed \cite{zhang2015optimizing,ma2017automatic,zhang2017improving,zeng2018framework,yang2019synetgy}. However, training deep neural networks on FPGA platform has not been investigated comprehensively.  Compared to inference, the training phase involves a much higher number of operations ($>$3X) with increased complexity~\cite{choi2018trainware}. 
The training phase also involves high intermediate data volume, necessitating high memory bandwidth and large storage. 
%Due to their high performance and parallelism, 
GPUs have been the de-facto for training tasks to meet immense computation requirements. However, GPUs' energy-efficiency is poor \cite{jouppi2017datacenter}, and they are not well-suited for on-device learning with limited power budget.   

To address this issue on the algorithm side, researchers have proposed low-precision training \cite{gupta2015deep,NIPS2017_6771}, frequency domain training \cite{ko2017design}, and sparse weight update~\cite{Sun2017ICML}. 
Techniques such as sparse weight update introduce irregular parallelism, making it more suitable for flexible FPGAs compared to GPUs \cite{nurvitadhi2017can}. FPGAs are well-suited for low-precision DNN algorithms as it provides large improvement in throughput and energy efficiency with low-precision arithmetic~\cite{wang2019deep}. To that end, implementing configurable training hardware on FPGA becomes crucial to exploit these algorithmic advances. % and allow on-device training on edge devices. 

On the hardware side, %training work is less explored due to the intricacy of training algorithms. 
several prior FPGA works have implemented training of fully-connected neural networks~\cite{liu2018fast,gomperts2011development,rafael2005fpga}. A floating-point FPGA accelerator~\cite{liu2017fpga} reported training of small CNNs using an uniform computation structure with a fixed number of multiply-and-accumulate (MAC) units. 
F-CNN~\cite{zhao2016f} presented a training framework where convolutions are done in FPGA and weight updates are performed in CPU.
TrainWare~\cite{choi2018trainware} implemented dedicated hardware for weight update using a fixed $N_{kx}$$\times$$N_{ky}$ MAC array as the local gradients window is reused only $N_{kx}$$\times$$N_{ky}$ times during weight gradient computation.  However, this is not suitable for FP/BP convolutions where there exists more kernel reuse.
%\cite{nurvitadhi2017can} provides a customizable PE architecture to support DNN matrix multiplications. 
%However, it does not consider loop optimizations like loop unrolling, loop tiling which is critical to balance the performance and required hardware resources. 
DeepTrain~\cite{kim2018deeptrain} presents an embedded platform for DNN training, but does not include back-propagation of pooling layers and DNN weight updates, which needs significant memory access.
%and do not present actual FPGA implementation or training results.
Overall, these works have not presented a 
%standalone end-to-end FPGA accelerator that can support various CNNs.
compiler-based FPGA accelerator that supports all phases of training for various CNNs.
Designing a standalone  FPGA accelerator for CNN training involves managing limited memory resources to support batch operations and 
%designing flexible training hardware to support 
different CNN configurations. % based on user-defined design variables and complexity in preprocessing/updating the weights during back-propagation. 

In this work, we propose a flexible FPGA accelerator that performs 
stochastic gradient descent (SGD) based training of various CNNs. 
%A library based CNN RTL compiler \cite{ma2017automatic,ma2016scalable,ma2017optimizing} is augmented to support training operations.
We extracted and designed training-specific operations and then developed a library based automatic RTL compiler to flexibly support training operations with different sizes of CNNs. 
The user provides the high-level CNN network configurations along with the design variables to characterize FPGA hardware usage to the RTL compiler. 
The RTL compiler generates a FPGA compatible training accelerator based on the user's requirements. The key contributions of this work are:
\begin{itemize}
 \item We present a comprehensive investigation of CNN training operations and challenges in FP, BP and WU stages. 
\item We developed a training-specific RTL module library and an RTL compiler to automatically implement CNN training accelerator with 16-bit fixed-point precision. %for design variables such as tile sizes, loop unroll factor, etc. 
%The library consists of parameterized Verilog modules to perform training-specific operations including upsampling and gradient computation.
\item A configurable FPGA hardware is presented for 
FP, BP and WU phases of the entire CNN training process using SGD with momentum. 
%An RTL compiler is developed to automatically generate FPGA-compatible training accelerator based on user's CNN model and resource specifications.
\item Our accelerator using Intel Stratix 10-GX FPGA is evaluated on training three different CNNs for CIFAR-10 dataset, achieving up to 479 GOPS of throughput.
\end{itemize}

\section{CNN Training Algorithm}

\begin{figure}[tb]
         \vskip -0.1in
        \includegraphics [trim={10cm 0 8.95cm 0},clip,width=0.77\columnwidth] {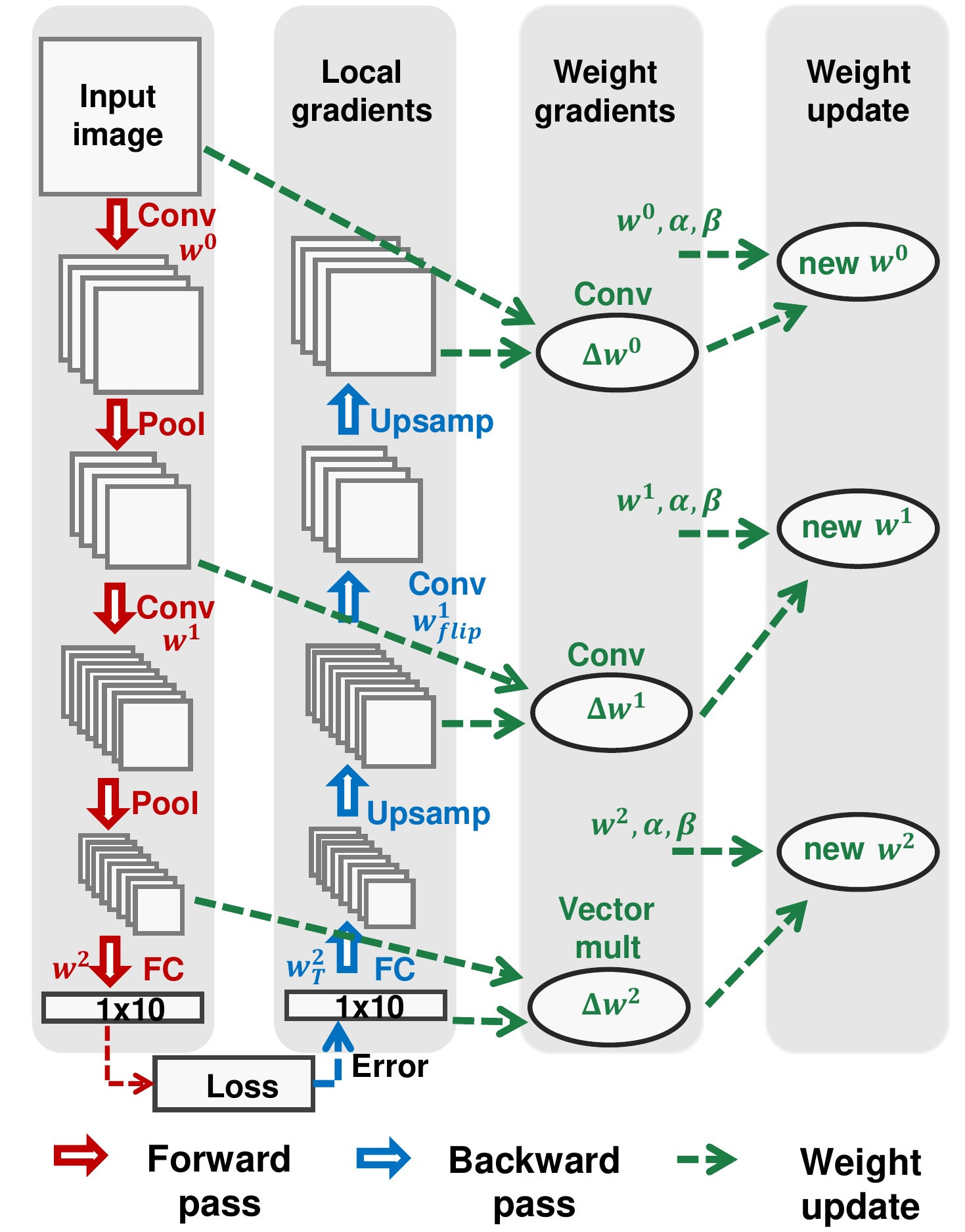} 
       \centering
        %\captionsetup{justification=centering}
       \vskip -0.01in
        \caption{SGD based CNN training dataflow illustrated for a simple 2C-2P-1FC model.} 
        \vskip -0.2in
        \label{train_algo}
\end{figure}

\begin{table}[b!]
\vskip -0.1in
\caption {CNN design variables}
\begin{tabular}{|c|c|c|c|}
\hline
 & \begin{tabular}[c]{@{}c@{}}Kernel size \\ width/height\end{tabular} & \begin{tabular}[c]{@{}c@{}}Output feature map\\ width/height/depth\end{tabular} & \begin{tabular}[c]{@{}c@{}}Input feature map\\ width/height/depth\end{tabular} \\ \hline
\begin{tabular}[c]{@{}c@{}}Convolution \\ dimensions\end{tabular} & \(N_{kx}\), \(N_{ky}\) & \(N_{ox}\), \(N_{oy}\), \(N_{of}\) & \(N_{ix}\), \(N_{iy}\), \(N_{if}\) \\ \hline
\begin{tabular}[c]{@{}c@{}}Loop unroll \\ factors\end{tabular} & \(P_{kx}\), \(P_{ky}\) & \(P_{ox}\), \(P_{oy}\), \(P_{of}\) & \(P_{ix}\), \(P_{iy}\), \(P_{if}\) \\ \hline
\end{tabular}
\label{cnn_Var}
\end{table}

%Back-propagation algorithm is extensively employed for CNN training. 
Fig.~\ref{train_algo} illustrates the dataflow of SGD based weight update for a simple 2C-2P-1FC CNN model. 
The CNN design variables and naming conventions %used throughout this paper 
are described in Table \ref{cnn_Var}.
Output activation value \(o_{x,y}^l\) %of a convolution layer pixel 
is given by Eq.~\eqref{outact}, where  \(w_{x,y}^l\) are kernel values and  \(a_{x,y}^{l-1}\) are activations from layer  \(l-1\).
\begin{equation}
\label{outact}
o_{x,y}^{l} = \sum_{x^\textnormal{\textquotesingle}}\sum_{y^\textnormal{\textquotesingle}} w_{x,y}^l a_{(x+x^\textnormal{\textquotesingle}),(y+y\textnormal{\textquotesingle})}^{l-1}
\end{equation}

In supervised training, each input is associated with a label. After the completion of the FP, the performance of the network is estimated using a cost function. Eq.~\eqref{costfunc} shows a quadratic cost function of output layer \(L\), where \(a_i\) is the obtained output value and \(y_i\) is the label. The derivative of the cost function with respect to output is also given in Eq.~\eqref{costfunc}.
\begin{equation}
\label{costfunc}
C=\frac{1}{2} \sum_{i}^L (a_i-y_i)^2,\quad \frac{\partial C}{\partial a_i^L}=(a_i-y_i)
\end{equation}
%\begin{equation}
%\label{costgrad}
%\frac{\partial C}{\partial a_i^L}=(a_i-y_i)
%\end{equation}

Error values are back-propagated to all hidden layers and the required deviation of weight parameters to minimize the error is calculated. The derivative of the cost function with respect to weight parameters provides the required deviation for the weight parameters \(\Delta w\) to minimize the error. By applying the basic chain rule, weight deviation \(\Delta w\) can be obtained by convolving the derivative of the cost function with layer output activations, which we term as local gradients and feedforward activations. Local gradients of layer \((l)\) can be obtained by convolving the gradients of the previous layer \((l-1)\) with its own convolution kernel. 

\begin{figure}[t!]
     \vskip -0.1in
       \includegraphics [trim={3.5cm 5.4cm 4.5cm 7.cm},clip,width=1 \columnwidth] {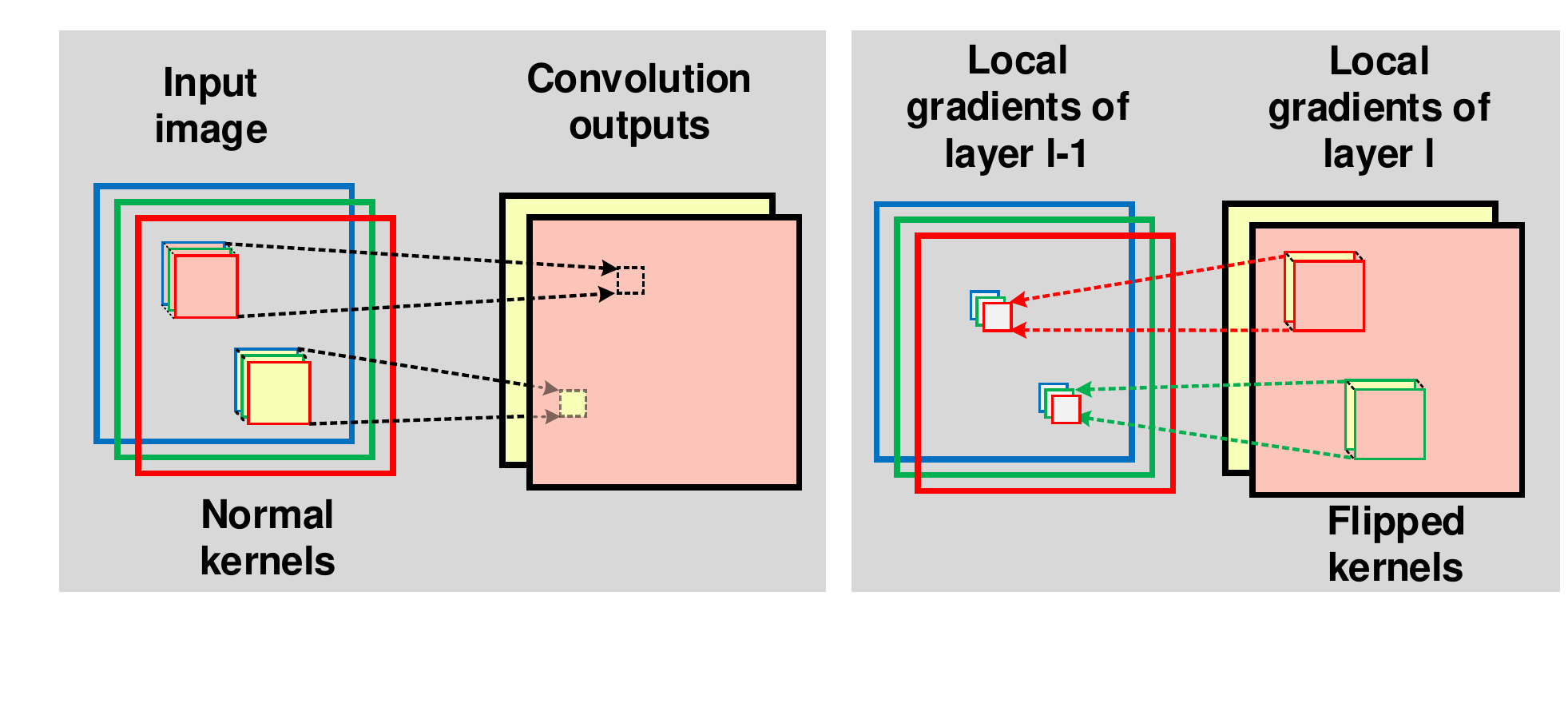}
       \centering
        %\captionsetup{justification=centering}
        \vskip -0.4in
         \subfloat[\label{fp} Feedforward convolutions.]{\hspace{.45\linewidth}}
        \subfloat[\label{bp} Backward convolutions.]{\hspace{.45\linewidth}}
        \vskip -0.02in
         \caption{Convolution operations and changes in kernels during FP and BP (\(N_{of}=2\), \(N_{if}=3\)).}
        \vskip -0.2in
        \label{kertrans}
\end{figure}

During these backward convolutions, the original kernel tensors are flipped. The differences of BP and FP convolutions are shown in Fig.~\ref{kertrans}. 
Fig.~\ref{fp} shows FP convolutions of input image with three input channels (\(N_{if}=3\)) and two sets of kernels to obtain two output feature maps (\(N_{of}=2\)). 
%Each output feature map depends on its kernel as shown in solid color. In each kernel, input channels have their association only same color bordered $N_{kx} \times N_{ky}$ kernel values. 
During BP, convolutions are performed using local gradients of previous layer and FP kernels, 
%FP kernels cannot be used directly as in BP, 
where the number of input channels and convolution depth are interchanged. In Fig.~\ref{bp}, it is shown that \(N_{if}=2\) and \(N_{of}=3\). Flipped kernels are used in BP convolutions to compute the local gradients.
\begin{equation}
\label{localgrad}
\delta_{x,y}^{l}=\varphi_l^\textnormal{\textquotesingle} (o_{x,y}^l) \sum_{x^\textnormal{\textquotesingle}}\sum_{y^\textnormal{\textquotesingle}} \delta_{x\textnormal{\textquotesingle},y\textnormal{\textquotesingle}}^{l+1} w_{(x-x^\textnormal{\textquotesingle}),(y-y\textnormal{\textquotesingle})}^{l+1}
\end{equation}
\vskip -0.1in
\begin{equation}
\label{wtgrad}
\Delta w_{n}=\frac{\partial C}{\partial w_{x,y}^L}=\sum_{x^\textnormal{\textquotesingle}}\sum_{y^\textnormal{\textquotesingle}} \delta_{x\textnormal{\textquotesingle},y\textnormal{\textquotesingle}}^{l} a_{(x+x^\textnormal{\textquotesingle}),(y+y\textnormal{\textquotesingle})}^{l-1}
\end{equation}
\vskip -0.1in
\begin{equation}
\label{wtupdate}
 w_{i,j}^l (n) = - \alpha \Delta w_{n} +  w_{i,j}^l (n-1)
\end{equation} 
\vskip -0.1in
\begin{equation}
\label{wtupdate_with_momentum}
 w_{i,j}^l (n) = \beta \Delta w_{n-1} - \alpha \Delta w_{n} +  w_{i,j}^l (n-1)
\end{equation} 

Local gradients of each layer \(l\) is computed using Eq.~\eqref{localgrad}, where \(w\) is the flipped kernel. Eq.~\eqref{wtgrad} is used for weight gradient computation, where \(l\) is local gradients of a layer and  \(\varphi_l^\textnormal{\textquotesingle} (x)\) is activation gradients of layer \(l\).  The weight gradients of layer \(l\) is obtained by the convolution of local gradient layer \(l\) and feedforward input activations of layer \(l\), involving large kernel sizes. One feature map of feedforward activation is convolved with one feature map of local gradients to obtain one kernel gradient (intra-tile accumulation). Hence, this weight gradient convolution results in a 4D output. These weight gradients are averaged over a batch and new weights are computed using gradient descent algorithm given by Eq.~\eqref{wtupdate}, where \(\alpha\) is learning rate, $w_{i,j}^l$($n-1$) is weights of previous batch and \( \Delta w_n\) is the average weight gradient. The weight update process can be accelerated by using past weight gradients as momentum. Eq.~\eqref{wtupdate_with_momentum} shows the weight update in SGD with momentum, where \(\beta\) is a hyper-parameter.

%\subsection{Training-Specific Computations During BP and WU}
The operations during BP are different to those of FP. In backward convolutions, the inputs are scaled by activation gradients, and convolutions are performed by applying 180-degree-rotated kernels. 
%In hardware implementation, the same kernels should be read in normal mode and transpose mode to support FP and BP operations. 
Similarly, fully-connected layers in BP also use transposed weight matrix to compute the local gradients. At the max-pooling node, the gradients propagate only through the selected maximum pixel location and all other pixels in the pooling window will be zero. Based on the pooling pixel index selected during FP, the gradients are upsampled and propagated back to the next layers.  

%The critical point to consider is that, 
During FP, we need to store not only the output activations, but also the activation gradients and max-pooling indices at all ReLU activations and max-pooling nodes. For ReLU, activation gradients are binary as the derivative of ReLU with respect to activations results in a step function. Our RTL library currently supports only ReLU activation function as it is less complex and widely used. During weight update of fully-connected layers, the weight gradients \(\Delta w\) are obtained by performing the outer product of the local gradient vector and the error vector. % to obtain a 2D weight gradient matrix. 
%This vector-vector multiplication is similar to element-wise multiplication and accumulation is not involved. 
%Special control signals need to be designed for the MAC units to support this operation. 
In convolution kernel updates, kernel gradient calculation involves convolution of input activations using local gradients as kernels, which are very large kernels.
%\textcolor{red}{Also, these convolutions require an additional outer loop such that each input feature map is considered as different images resulting in intra-tile accumulation.}
Each of these convolutions is considered as an FP convolution with \(N_{if}=1\) and results in \(N_{of}\) kernel gradients. 
%Actual \(N_{if}\) local gradients are considered as different images \(N_{im}\).
%Convolutions are repeated \(N_{im}\) times to obtain all \(N_{if} \times N_{of}\) kernel gradients. 
To reuse FP convolution control logic, we employed an additional outer loop to iterate through the actual \(N_{if}\) local gradients.
%required to iterate through \(N_{im}\).

%\subsection{CNN Training Using Fixed-Point Precision}
Unlike CNN inference, CNN training usually requires higher precision. In this work, weights, activations, and local/weight gradients are represented with 16-bit fixed-point precision to ensure good training accuracy~\cite{chen2017fxpnet, gupta2015deep}. 
Compared to floating-point precision, fixed-point precision training leads to more energy-efficient FPGA design, but requires more dedicated resolution/range assignment for different variables. 
%\textcolor{red}{
%Based on heuristics, we allocate the number of integer bits to be 4 times of the initial range of weights in each layer to avoid overflow of weights during training.}

\section{CNN Training Hardware}
\subsection{RTL Compiler and Algorithm Mapping}
\begin{figure}[tb]
       %\vskip -0.1in
       \includegraphics [trim={2.33cm 3.3cm 2.9cm 3.3cm},clip,width=1 \columnwidth] {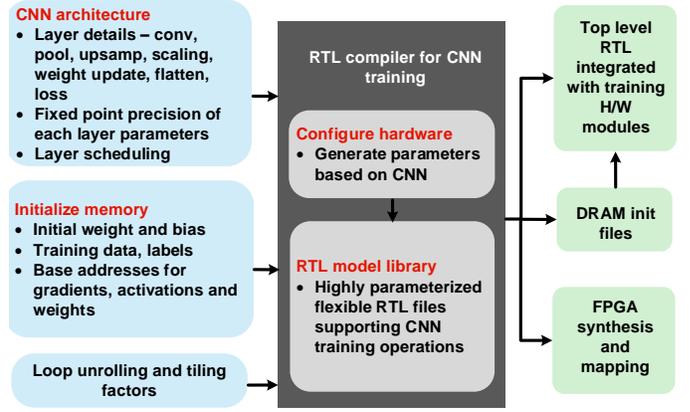}
       \centering
        %\captionsetup{justification=centering}
        \vskip -0.1in
        \caption{Proposed RTL compiler automatically generates FPGA training accelerator from high-level CNN description.} 
        \label{rtl_comp}
        \vskip -0.1in
\end{figure}

To map various CNN algorithms with user defined hardware constraints onto FPGA, an RTL compiler for CNN training was developed. Fig.~\ref{rtl_comp} shows the compiler tool flow from high-level CNN description to CNN training accelerator. According to the operations in each layer and FPGA design parameters (e.g. unroll and tiling factors), optimized handwritten Verilog modules are chosen from the RTL library to automatically generate a CNN training accelerator. The RTL library consists of Verilog modules that are specially designed to support training operations. Only the selected modules from the RTL library based on the training algorithm will be synthesized. Execution of training operations in one iteration of a batch can be scheduled sequentially similar to layer-by-layer execution of inference tasks. Each training image in a batch is processed sequentially. The scheduling of layer execution is done using the RTL compiler, and control logic parameters are generated. 

\subsection{Training Accelerator Architecture}

\begin{figure}[tb]
       %\vskip -0.1in
       \includegraphics [trim={1.5cm 0.2cm 1.5cm 0.2cm},clip,width=1 \columnwidth] {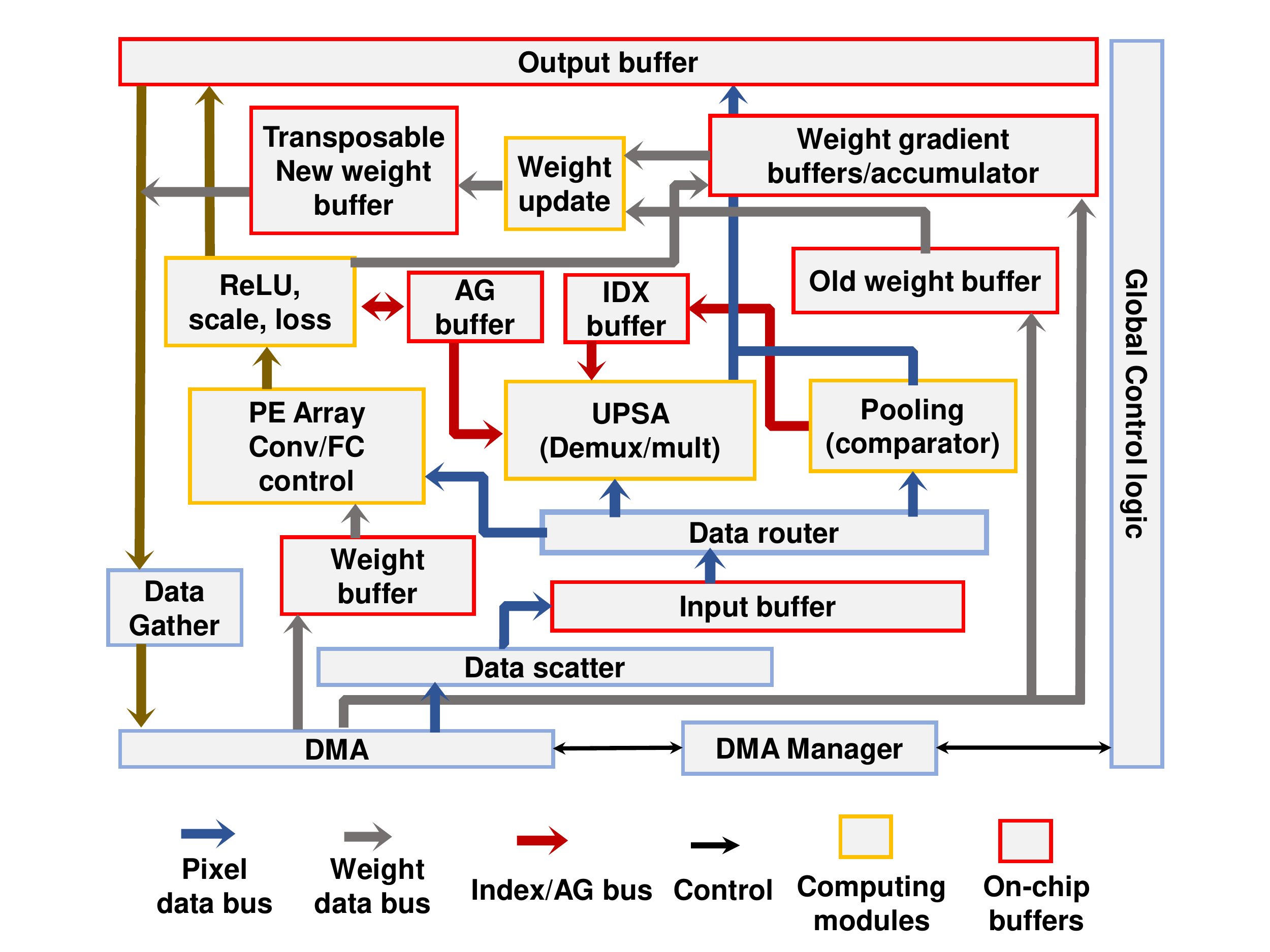}
       \centering
        %\captionsetup{justification=centering}
        \vskip -0.05in
        \caption{Top-level block diagram of CNN training accelerator.} 
        \label{block_diagram}
        \vskip -0.1in
\end{figure}

Fig.~\ref{block_diagram} shows the top-level diagram and dataflow of the CNN training accelerator. % based on the stochastic gradient descent algorithm. 
The global control logic governs all modules to ensure proper CNN functionalities with layer-by-layer computation, and is controlled by the parameters generated by the RTL compiler.
DRAM stores all the initial weight parameters, intermediate activations and computed weight/loss gradients using 16-bit fixed-point precision. DMA control generates the required DMA descriptors based on the layer type and tile sizes to read from and write to DRAM. 
A tile is a portion of data stored in on-chip buffers after/before reading/writing back to DRAM.
%The acceleration process is triggered once all the training images and initial parameters are loaded to DRAM. 
Convolution, max-pooling and upsampling operations are considered as \textit{key layers}, and ReLU, flatten, loss unit, and scaling unit are referred to as \textit{affiliated layers}. 
Key layers read new data from DRAM and affiliated layers use outputs of key layers.

%In FP/BP convolutions, transposable weights are read from DRAM and read in normal/transpose mode. 
On-chip buffers store activation gradients and max-pooling indices. The pooling window size (e.g. 2x2) determines the bitwidth of max-pooling indices (e.g. 2-bit). 
After FP, loss is computed using outputs and labels. Our RTL library currently supports square hinge loss and euclidean loss functions, and this can be easily expanded to support other loss functions. 
%Implementation of the configurable upsampling unit is done using highly parameterized Verilog scripts and same hardware is reused for various CNN design parameters. 
%Upsampling unit is enabled during the backward pass. Scaling unit reads the appropriate activation gradients from on-chip BRAMs and scales the output activations. 
Data scatter and data gather modules are used to convert the DRAM storage pattern to on-chip buffer storage pattern and vice versa. Data router reads the data from input buffers and routes it to the selected key layer according to the array sizes. Weight update unit and weight gradient buffers are used to compute new weights based on SGD with momentum.

\begin{figure}[tb]
      %\vskip -0.1in
       \includegraphics [trim={2.2cm 0.2cm 2.5cm 2.4cm},clip,width=1 \columnwidth] {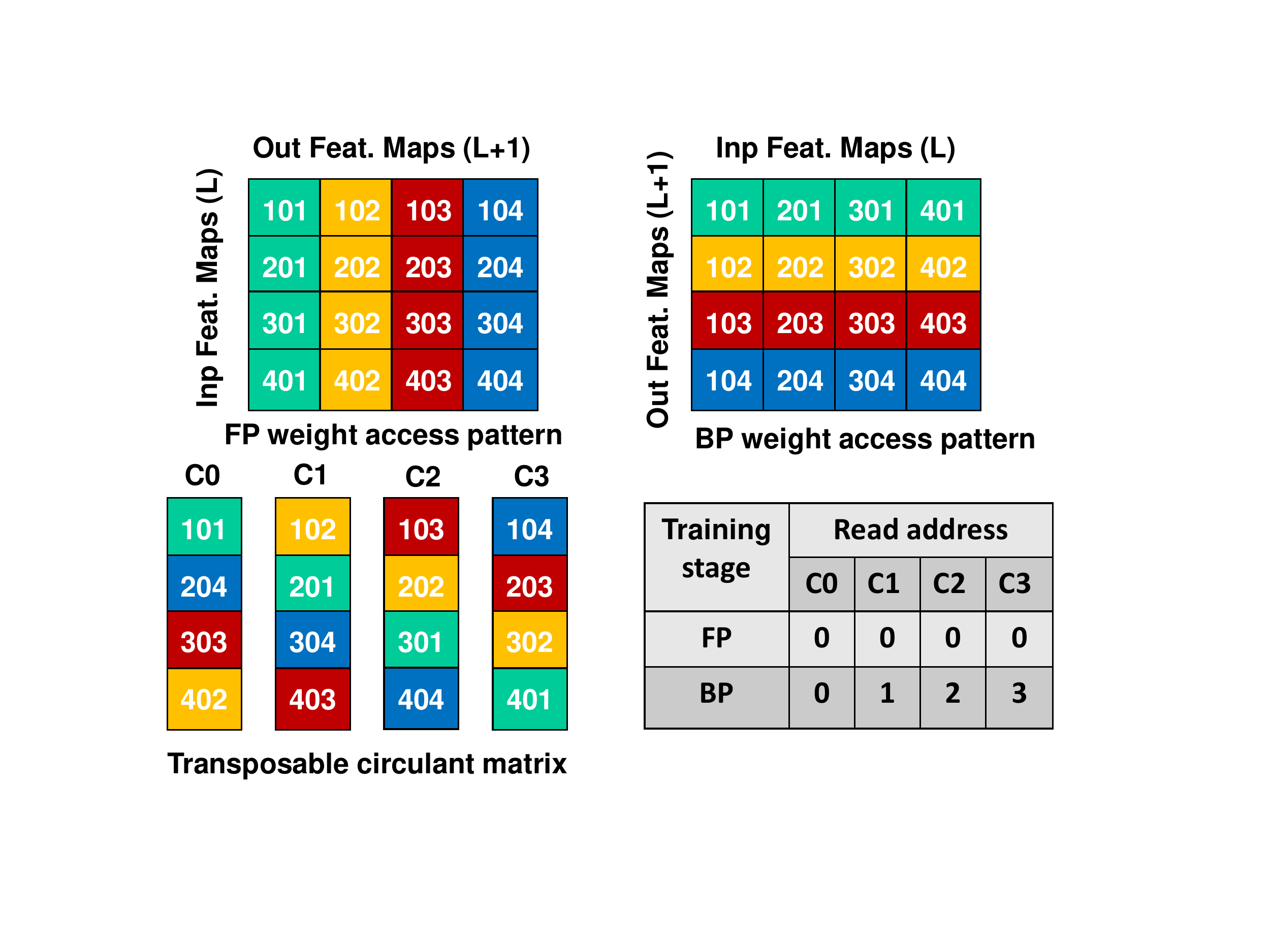}
       \centering
        %\captionsetup{justification=centering}
        \vskip -0.5in
        \caption{Proposed transposable weight buffer stores weights in a circulant matrix, enabling both normal and transpose read.} 
        \label{transbuf}
        \vskip -0.2in
\end{figure}

\subsection{MAC array}

\begin{figure}[tb]
      %\vskip -0.1in
       \includegraphics [trim={2.6cm 0.5cm 4cm 0.5cm},clip,width=1 \columnwidth] {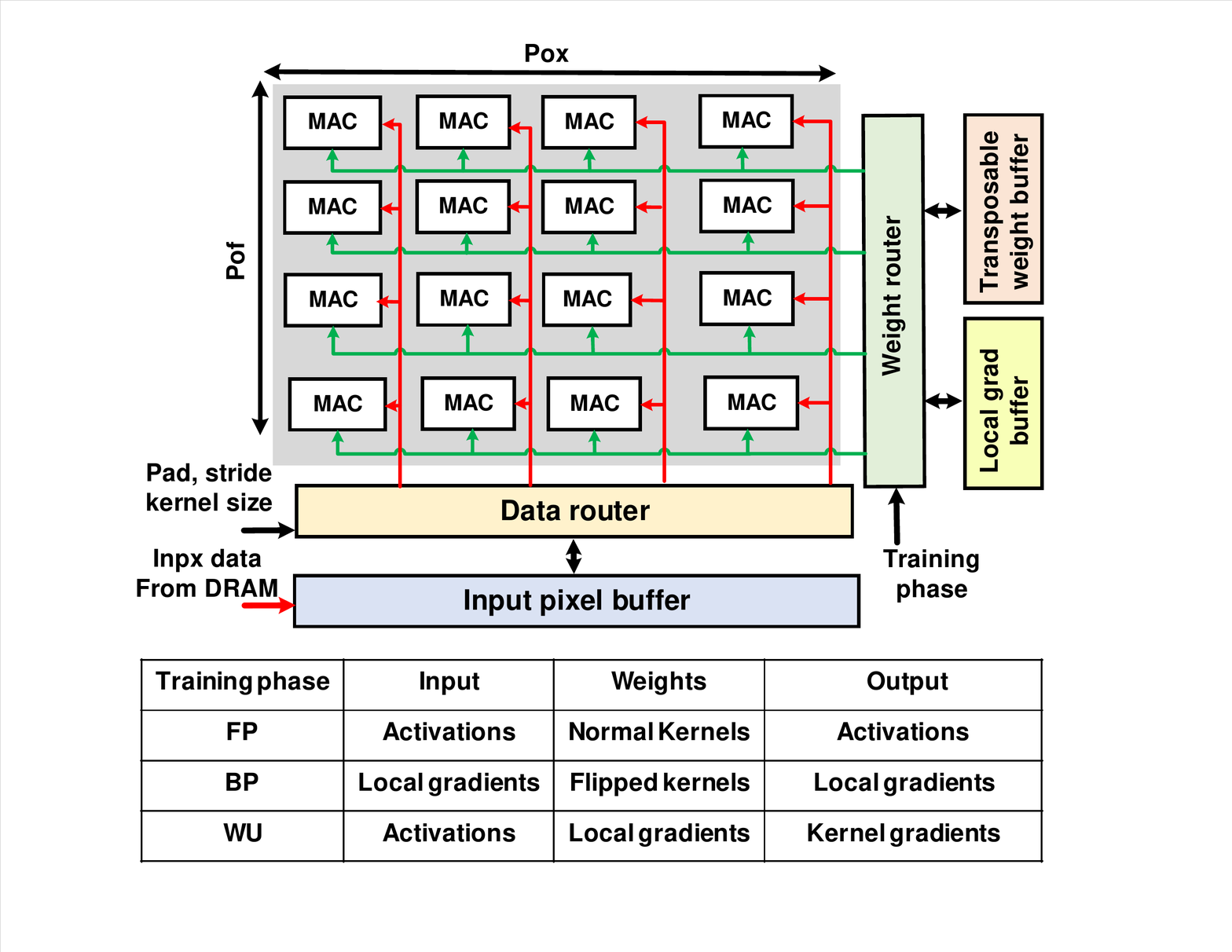}
       \centering
        %\captionsetup{justification=centering}
        \vskip -0.2in
        \caption{Systolic MAC array is reused for training phases of FP, BP and WU, by feeding different activations/gradients/kernels.}
        \label{macarch}
        \vskip -0.1in
\end{figure}

Fig.~\ref{macarch} shows the 2D systolic MAC array used for the training accelerator. MAC array size is determined by the RTL compiler based on the loop unroll factors \(P_{ox}, P_{oy}, P_{of}\). In Fig.~\ref{macarch}, each MAC row has a different set of weights but share the same input feature map data computing \(P_{of}\) output pixels. Each column shares the same weights, but different input data computing \(P_{ox}\) or \(P_{oy}\) output pixels in parallel. Data router reads the input data and routes it to MAC units considering pad and stride sizes of the layer. Weight router distributes weights or local gradients based on the training phase. Table in Fig.~\ref{macarch} summarizes how the MAC array is reused with different inputs/outputs for training phases of FP, BP and WU.

%\begin{table}[b]
%  \centering
%\begin{tabular}{|c|c|c|c|}
%\hline
%\begin{tabular}[c]{@{}c@{}}Training\\   Phase\end{tabular} & Input & Weights & Output \\ %\hline
%FP & Activations & Normal kernels & Activations \\ \hline
%BP & Local gradients & Flipped kernels & Local gradients \\ \hline
%WU & Activations & Local gradients & kernel gradients \\ \hline
%\end{tabular}
%\caption{MAC array inputs and outputs during different training phases}
%\label{mac_tabble}
%\end{table}
 \vskip -0.4in
\subsection{Transposable Weight Buffer}
BP involves convolution of flipped kernels and the local gradients. %, to back-propagate the error to next layers. 
Therefore, every convolution kernel is used twice in one iteration: 1) normal weights are applied during FP, and 2) rotated weights are used in BP (Fig.~\ref{kertrans}).
To achieve this without duplicating kernel storage, the kernels are stored in special transposable buffers that we propose, where data can be read both in non-transpose and transpose modes. As shown in Fig.~\ref{transbuf}, the proposed transposable buffer stores the kernels in the form of a circulant matrix using column buffers. For 2D kernels, each \(N_{kx} \times N_{ky}\) kernel is considered as one block and each row has \(P_{of}\) blocks of kernels, where \(P_{of}\) represents the number of output feature maps that can be computed in parallel. %All the input feature map kernels \((N_{if})\) out of \(P_{of}\) output feature maps are stored one after the other. 
%Kernels corresponding to the next \(P_{of}\) output feature maps are stored subsequently in the same manner. 
During backward convolution, not only the kernel is rotated by 180 degrees but also the input and output feature maps will be interchanged. In the proposed transposable buffer, every row of kernel blocks is circularly rotated and stored in the form of a circulant matrix in the single-port column buffers (Fig.~\ref{transbuf}). In the non-transpose mode, each column buffer shares the same read address, and in transpose mode, each column buffer obtains shifted addresses from the address translator unit. Address translator generates read/write addresses for column buffers for every transposable block. In each transposable block, the address vectors and the data are circularly shifted using shift registers.
 \vskip -0.2in
\subsection{Weight Update Unit}
\begin{figure}[tb]
       %\vskip -0.125in
       \includegraphics [trim={3.2cm 3.5cm 3.9cm 3cm},clip,width=0.95\columnwidth] {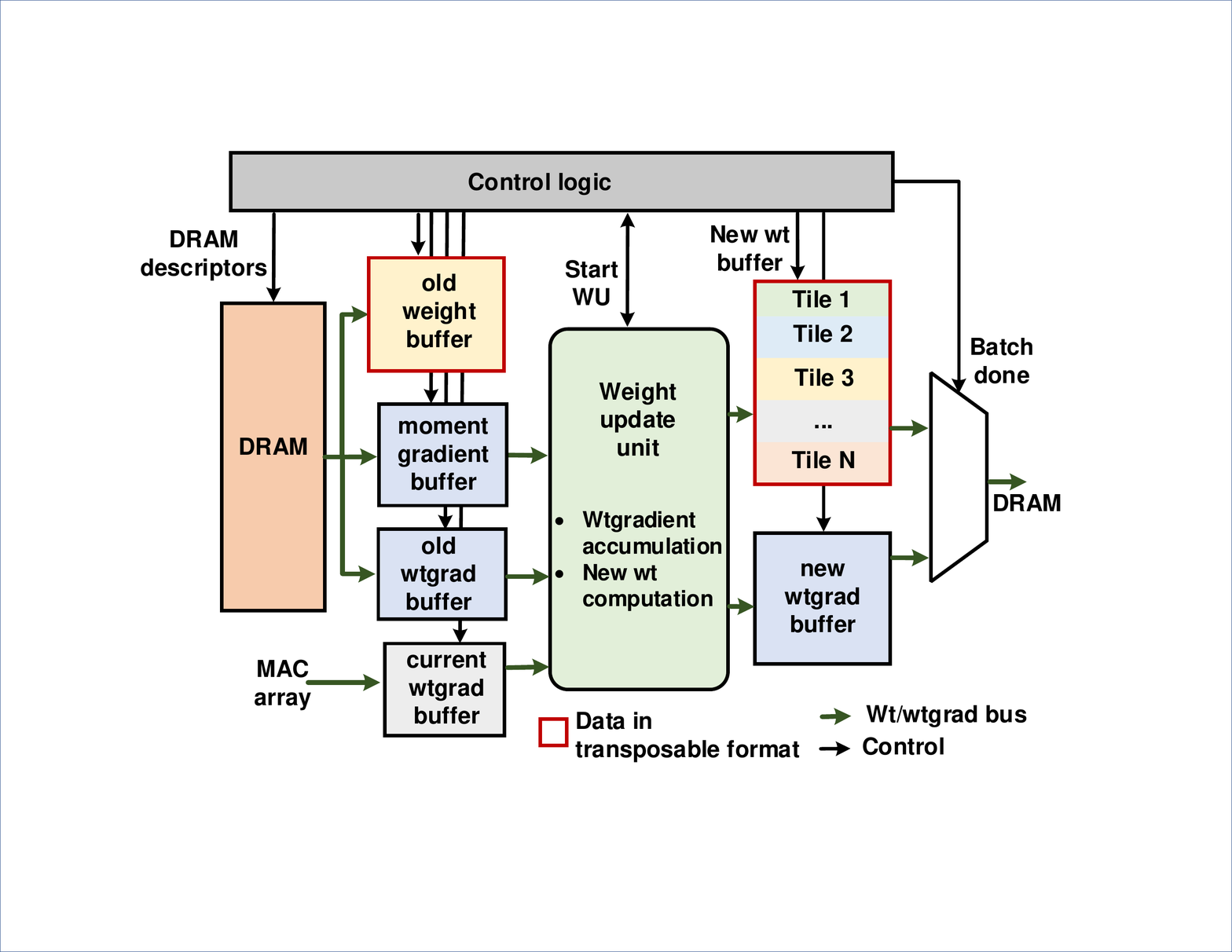}
       \centering
        %\captionsetup{justification=centering}
        \vskip -0.15in
        \caption{Block diagram of weight update unit.} 
        \label{wu}
        \vskip -0.20in
\end{figure}

Weight gradients are calculated by convolving the feedforward activations with the local gradients. Convolution control logic is configurable to support tile-by-tile computation, intra-tile accumulation and  large kernel sizes needed for weight gradient computation. Fig.~\ref{wu} shows the dataflow after the computation of weight gradients. For every new training image in a batch, newly computed weight gradients are accumulated with old weight gradients.
This accumulation is done tile-by-tile %for efficient utilization of on-chip buffers. This process is 
and repeated for the entire batch of images while the accumulated gradients are stored in DRAM. At the end of the batch, as the weight gradients get accumulated, old weights and past weight gradients are also read from DRAM, and new weights are computed following Eq.~\eqref{wtupdate_with_momentum}. 

Weights are initially stored in transposable format in DRAM as aforementioned. The entire transposable weights of layer \(l\) are read from DRAM to the old weight buffer. New weights are computed tile-by-tile and written back in transposable format to the new weight buffer. After completing the last tile's computation, the new weights are written back to DRAM. Control logic translates the address for transposable read/write operations, generates DRAM descriptors according to tile count and generates addresses to read newly computed weight gradients. %in weight format. 
Fully-connected weight update follows the same dataflow, but gradients are computed by outer product of local gradient vector and activation vector. %Constant learning rate is applied throughout the training process, and 
16-bit fixed-point precision is used for all weights and gradient computation.

%\subsection{Bias Update Unit}
%
%Bias update in a layer is computed similarly to weight updates. Bias gradients for convolution %layers are computed by taking the mean of the local gradients over each feature map. Local %gradients of a convolution layer can come from either an upsampling unit or a convolution %unit. The global control logic appropriately selects the source of local gradients based on %the CNN model structure. Bias update unit performs the accumulation of selected local %gradients. On-chip buffers stores the bias and bias gradients, and bias update computations %are performed at the end of the batch.

\subsection{Efficient MAC Usage in Weight Update layers}

 During FP and BP, the MAC array is designed to compute convolutions for $P_{ox}$$\times$$P_{oy}$$\times$$P_{of}$ pixels in parallel. 
 %$P_{ox}$, $P_{oy}$, $P_{of}$ are user-defined loop unroll factors. 
 Regarding convolutions required for weight updates, however, the output feature map size $N_{ox}$, $N_{oy}$ is less as the outputs are kernel gradients. This results in inefficient usage of MAC units, since most of them will be idle. It also consumes more output buffer storage in order to store $P_{ox}$$\times$$P_{oy}$$\times$$P_{of}$ block of output data. To overcome this, MAC load balance unit was designed to utilize the idle MAC units.

 \begin{figure}[tb]
       \includegraphics [trim={3cm 3cm 3cm 1cm},clip,width=0.9\columnwidth] {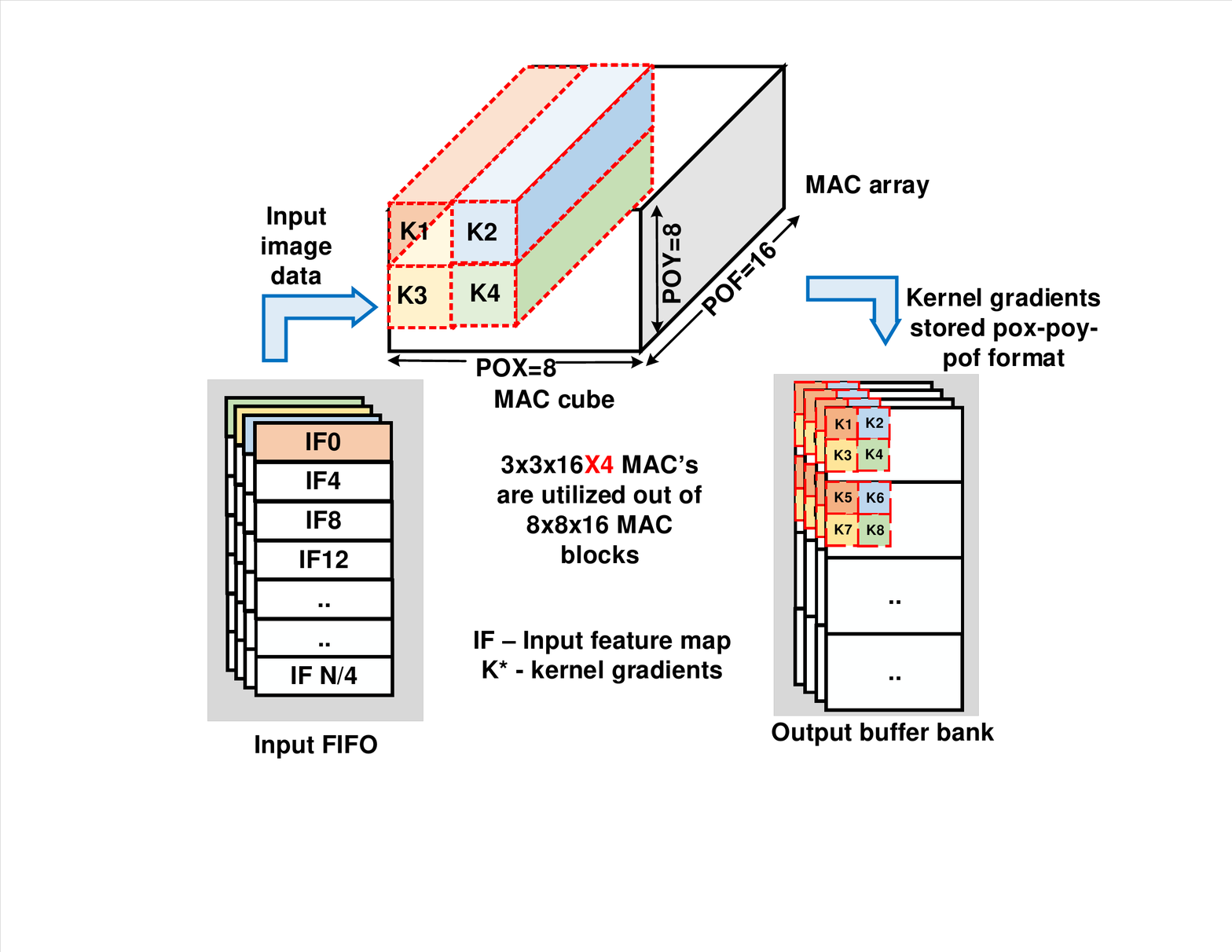}
       \centering
        %\captionsetup{justification=centering}
        \vskip -0.15in
        \caption{Operation of MAC load balancing unit during convolution weight gradient computation.} 
        \label{load_balance}
        \vskip -0.1in
\end{figure}

 The MAC load balance unit employs additional input buffers to feed the data to the MAC units in parallel. If buffer usage is critical, this optimization can be disabled by the RTL compiler. %to generate new accelerator. 
 Fig.~\ref{load_balance} shows the operation of MAC load balancing unit, when $P_{ox}$=8, $P_{oy}$=8, $P_{of}$=16 and kernel size is $N_{ox}$=3, $N_{oy}$=3, $N_{of}$=16. In this example, four kernel gradients are computed in parallel, reducing the latency by 4X without additional MAC units. The output buffer is also efficiently used.
 
\subsection{Upsampling and Scaling module}

%\begin{figure}[tb]
%       %\vskip -0.10in
%       \includegraphics [trim={0.7cm 0.6cm 0.65cm 0.6cm},clip,width=\columnwidth] {Upsampler.pdf}
%       \centering
%        \caption{The upsampling unit receives input from index buffer, activation gradient buffer and input buffer.} 
%        \vskip -0.2in
%        \label{upsamp1}
%\end{figure}
During BP, the local gradient at the max-pooling node is propagated to convolution layers only through the maximum pixel position selected in FP. The gradients of unselected pixels are zero, as they do not contribute to the error. If the max-pooling unit receives the input from ReLU node, then the upsampled gradients should also be scaled by the feedforward activation gradients to compute the gradients of ReLU node.  

%Fig.~\ref{upsamp1} shows the upsampling unit and input/output buffers. %along with index/activation gradient buffers. 
%Index buffer storage pattern follows the same output storage pattern of max-pooling layers.
\vskip -0.2in
During FP, max-pooling indices are stored tile-by-tile inside the on-chip index buffers. 
%Activation gradients are stored similarly to convolution layers and stored on-chip. The storage pattern of index and activation gradients are not altered to reuse the write controller of output buffers. 
Each layer has its own index and activation gradient buffers. The local gradients computed in the previous iteration is read from DRAM and stored in input buffers. Data router unit rearranges the data of index, input and activation gradient buffers and sends it to the upsampling unit. 
Each %processing element of the 
upsampling unit consists of a demultiplexer and a multiplier unit. The gradient is conveyed as the demultiplexer input and the index serves as the select signal. For pooling window size of $k$, each processing block produces $k$$\times$$k$ pixel data corresponding to $k$ rows of the output feature map. After each operation, $k$ rows of activation gradients are read and the demultiplexer outputs are scaled. 
%Scaling during the upsampling stage is optional and can be skipped by tuning the parameters.

\section{Results}

\subsection{Experimental Setup}

\begin{table*}[tb]
\caption{Evaluation of CNN training accelerator on Stratix 10 FPGA , using 16-bit fixed point precision. CIFAR10-1X refers to network structure of 16C3-16C3-P-32C3-32C3-P-64C3-64C3-P-FC, and 2X/4X designs refer to accordingly wider CNNs.}
\begin{tabular}{|c|c|c|c|c|c|c|c|c|c|c|c|c|}
\hline
\multirow{2}{*}{CNN network} & \multicolumn{3}{c|}{Resource} & \multicolumn{5}{c|}{Power (W)} & \multicolumn{3}{c|}{Latency per epoch (s)} & Throughput \\ \cline{2-12}
 & DSP & ALM & BRAM & DSP & RAM & Logic & clock & Pstatic & BS-10 & BS-20 & BS-40 & GOPs \\ \hline
CIFAR-10 1X & 1699 (30\%) & 20.8K (19\%) & 10.6(4.4\%) & 0.58 & 5.7 & 2.4 & 1.68 & 10.28 & 18.19 & 18.07 & 18.01 & 163 \\ \hline
CIFAR-10 2X & 3363 (58\%) & 415K (44\%) & 22.8(9.5\%) & 1.05  &11.2  & 6.6 & 2.97 & 11 & 41.7 & 41.30 & 41 & 282 \\ \hline
CIFAR-10 4X & 5760(100\%) & 720K(76.2\%)  & 54.5(22.4\%) & 3.48 & 14.6 & 11 & 4.95 & 16.47 & 98.2 & 96.87 & 96.18 & 479 \\ \hline
\end{tabular}
\label{restable}
\end{table*}

The FPGA accelerator generated by the compiler was synthesized using Intel Quartus 17.1 at 240MHz frequency. We used Stratix 10 GX FPGA as the target hardware, which includes 240 Mbits of BRAM, 5,760 DSP blocks, and 93K ALMs. The development kit~\cite{devkit} is equipped with 4Gb DDR3 DRAM with 16.9Gb/s bandwidth. 
Weights, weight/local gradients, and activations use 16-bit fixed point precision. We trained representative CNNs for CIFAR-10 dataset. `1X' CNN has the structure of 16C3-16C3-P-32C3-32C3-P-64C3-64C3-P-FC. 2X and 4X CNN models exhibit 2X and 4X more input/output feature maps for each layer, and could achieve higher accuracy. Unroll factor of 8 was used for output image $x$ and $y$ dimensions. For output feature maps, 16, 32, 64 was used as unroll factors for 1X, 2X and 4X CNNs, resulting in 8x8x16 (1,024), 8x8x32 (2,048), 8x8x64 (4,096) MAC arrays, respectively. Batch size (BS) of up to 40 and learning rate of 0.002 was used for training. %, towards achieving better accuracy. 
Latency was measured using simulation of the synthesized accelerator. DRAM modules and Intel IPs were used in the testbench adhering to DRAM protocols. We also developed a custom fixed-point precision training model using PyTorch  \cite{paszke2017automatic} to verify the functionality of the FPGA design with the same precision.
%to obtain the software results running on Nvidia Titan XP GPU.

\subsection{Results and Analysis}

\begin{figure}[b!]
       \vskip -0.2in
       \includegraphics [trim={3.05cm 6.3cm 4cm 5.2cm},clip,width=0.9\columnwidth] {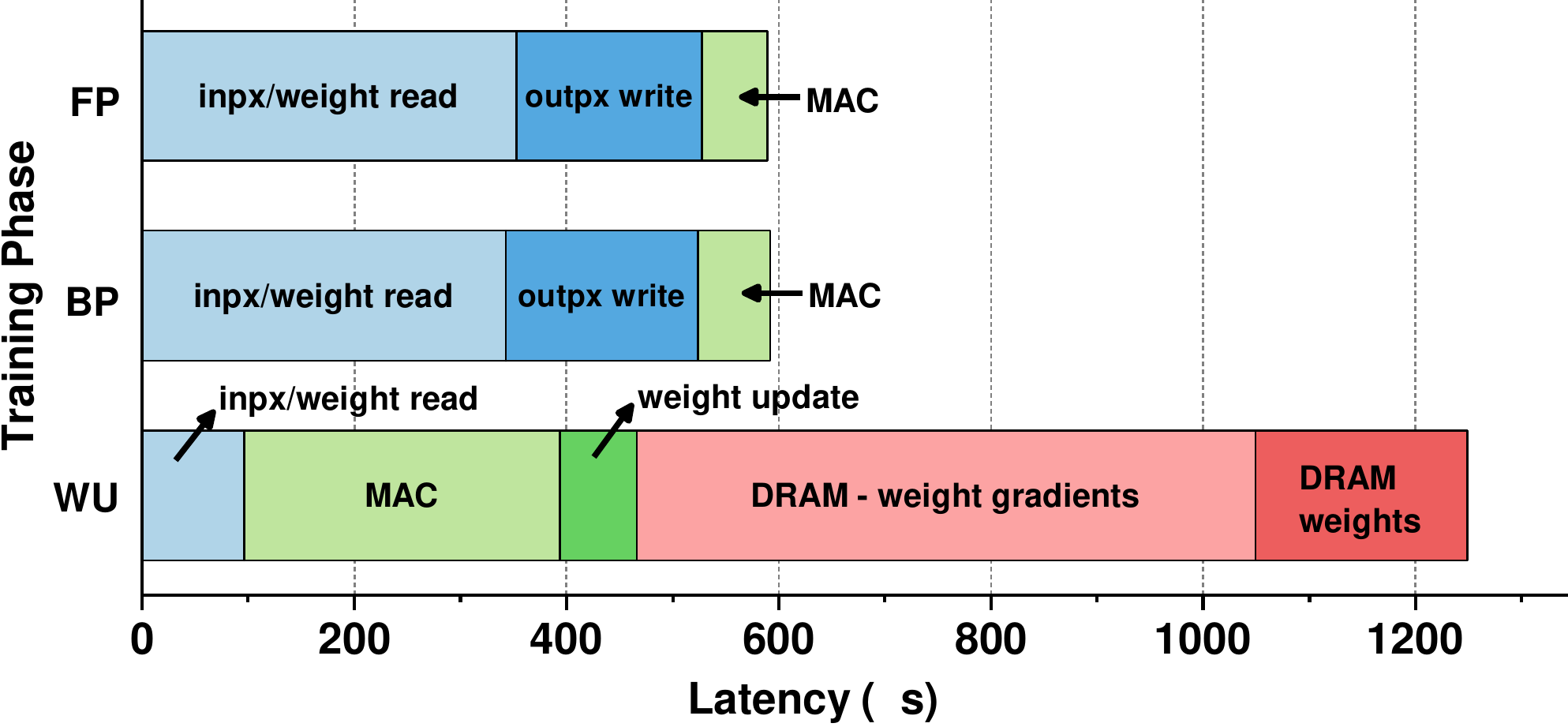}
       \centering
        %\captionsetup{justification=centering}
        %\vskip -0.015in
        \caption{Latency breakdown of CIFAR-10 4X CNN for FP, BP and WU for the last iteration of a batch.} 
        \label{lb}
\end{figure}

Table \ref{restable} shows the comparison of CNN training performance and resource utilization for three different CNNs for CIFAR-10 dataset. The FPGA accelerator was generated from the RTL compiler using high-level description of training parameters and design variables. FPGA power numbers are obtained after routing stage from Quartus power analyzer and Intel Early Power Estimator tools using the data toggling activity from functional simulation at the junction temperature of 65\textdegree{}C. Tiling of activations and weight gradients greatly reduces the on chip buffer usage. BRAM utilization is low because of the tiling and size of the intermediate activations and number of parameters. Training of each image in a batch is done sequentially,  larger batch sizes results in less number of weight updates in one epoch resulting in improvement in latency.

\begin{table}[t]
\vskip -0.1in
\caption{Performance comparison with GPU.}
\begin{tabular}{|c|c|c|c|c|c|c|}
\hline
 & \multicolumn{3}{c|}{Throughput (GOPs)} & \multicolumn{3}{c|}{Efficiency (GOPs/W)} \\ \hline
Device & \multicolumn{2}{c|}{Titan XP} & FPGA & \multicolumn{2}{c|}{Titan XP} & FPGA \\ \hline
Batch size & 1 & 40 & 1/40 & 1 & 40 & 1/40 \\ \hline
CIFAR-10 1X & 45.67 & 551.87 & 163 & 0.50 & 3.68 & 7.90 \\ \hline
CIFAR-10 2X & 128.84 & 1337.98 & 282 & 1.30 & 8.26 & 8.59 \\ \hline
CIFAR-10 4X & 331.41 & 2353.79 & 479 & 2.91 & 13.45 & 9.49 \\ \hline
\end{tabular}
\vskip -0.15in
\label{gpu_table}
\end{table}

Performance comparison of our accelerator implementation on Stratix 10 FPGA and Titan XP GPU is shown in Table \ref{gpu_table}. Our performance remains the same for different batch sizes as the images in a batch are processed sequentially one after the other. Our implementation shows better energy efficiency for smaller batch sizes. For batch size of 40, the 4X model shows less energy-efficiency than GPU, due to limited DRAM bandwidth (30X less than Titan XP). Stable and reliable training can also be achieved with smaller batch sizes as it provides more up-to-date gradient calculations \cite{masters2018revisiting}.

\begin{figure}[tb]
        \vskip -0.2in
        \includegraphics [trim={2.8cm 6cm 4cm 5cm},clip,width=1 \columnwidth] {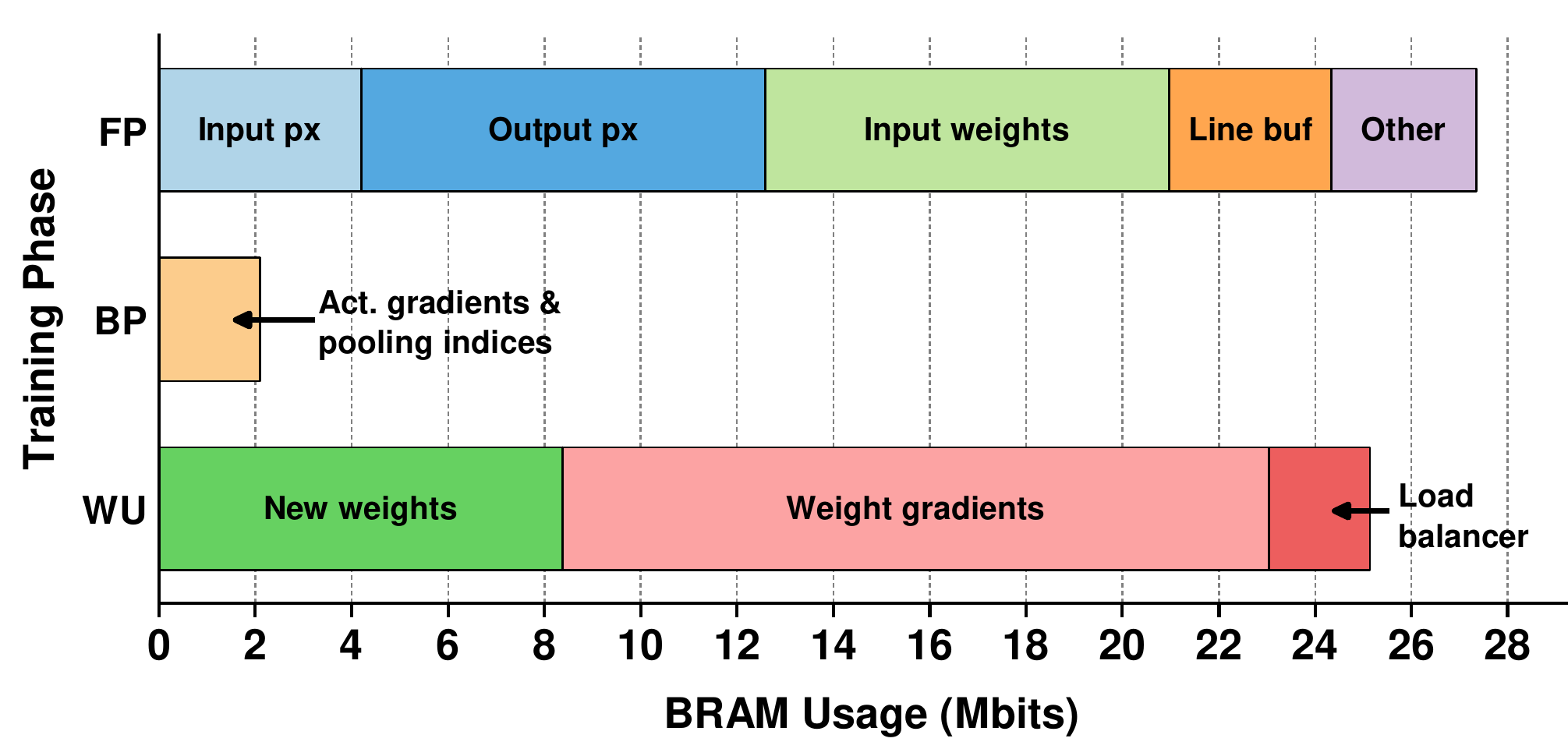}
       \centering
        %\captionsetup{justification=centering}
        \caption{Buffer usage breakdown of CIFAR-10 4X CNN.}
        \label{bb}
         \vskip -0.2in
\end{figure}

To flexibly support arbitrary sizes of CNNs, all intermediate outputs are stored in DRAM. Fig.~\ref{lb} shows the latency breakdown %in each layer
during different stages of training. Weight update layers will have large DRAM access latency due to access of past weight gradients, weights and storing back the updated values. 51\% percent of the overall latency in one iteration of a batch is consumed in weight update layers. By sacrificing the flexibility of the hardware, this latency could be significantly reduced by using on-chip buffers for weight/gradient storage. 
%The latency of the weight update layers depends on the parameters associated with it. 

Old weight gradients are read from DRAM tile-by-tile during computation of current weight gradients.
%\sout{wherever logic latency is more than the memory access latency.} 
Double buffering scheme is employed to hide the memory access latency 
%by using the compute time to read next tile data 
\cite {zhang2015optimizing},
which reduced the latency of weight update layers by 11\%.
%compared to baseline design by using the double buffering scheme.
The logic latency in weight update layers is reduced by 4X, using the load balancing technique for MAC arrays. Logic in weight update layers refer to convolution operations to generate weight gradients and weight update is referred to computation of new weights. Tile sizes are carefully chosen to efficiently map compute-/memory-bounded layers. All buffers can be controlled by tile sizes apart from weight buffers, where the entire weights are read from transposable DRAM. 

Fig.~\ref{bb} shows the breakdown of buffer utilization for three different phases of training. The weight buffer size is decided by the largest layer weights. Double buffering technique is used for all other buffers, thereby hiding DRAM latency. %Training performance of the network on CIFAR-10 dataset in fixed point is shown in Fig.~\ref{epoch_loss}. 
The 1X design achieves 73\% CIFAR-10 accuracy at 50 epochs with learning rate of 0.002 and batch size of 40 (similar to baseline with floating-point precision). Higher accuracy will be achievable %\sout{with longer training time and deeper/wider CNNs.} 
with addition of integer batch normalization and adaptive fixed point features~\cite{chen2017fxpnet} to our RTL module library.

%\begin{figure}[h!]
%       \includegraphics [trim={6cm 3cm 6cm 3cm},clip,width=1 \columnwidth] %{epoch_loss}
%       \centering
%        \captionsetup{justification=centering}
%        \vskip -0.05in
%        \caption{ (NOT THE REAL PLOT)Variation of loss after each epoch using fixed %point training with batch size of XX} 
%        \label{epoch_loss}
%        \vskip -0.05in
%\end{figure}

\section{Conclusion}
In this paper, we presented an automatic RTL compiler based end-to-end CNN training accelerator. 
CNN training operations are implemented by optimized and parameterized custom Verilog modules, and the accelerator is flexible to support various FPGA design parameters. The training performance is evaluated on Intel Stratix-10 GX FPGA for three different CNNs for CIFAR-10 dataset. The proposed training accelerator achieves throughput of up to 479 GOPS at 240MHz for CNNs with 2M parameters. 
%16-bit fixed point training achieved classification accuracy of 73\% on CIFAR-10 dataset.

\bibliographystyle{ieeetr}
\bibliography{references}
\include{references}

\end{document}